\documentclass{article}
\usepackage{spconf,amsmath,graphicx}
\usepackage{wrapfig}
\usepackage{subfigure}
\usepackage{braket}
\usepackage{multirow}
\usepackage{bigstrut}
\usepackage{booktabs}
\usepackage{svg}
\usepackage{enumitem}
\setlist{nosep, leftmargin=14pt}

\usepackage{mwe} 

\title{Label uncertainty-guided multi-stream model for disease screening}
\name{Chi Liu$^{1}$ \qquad Zongyuan Ge$^{2}$ \qquad Mingguang He$^{1,3}$ \qquad Xiaotong Han$^{1}$ \sthanks{indicates the corresponding author. Email: lh.201205@aliyun.com}}

\address{$^{1}$ Zhongshan Ophthalmic Center, Sun Yat-sen University, Guangzhou, China \\ 
$^{2}$ Monash eResearch Centre, Monash University, Melbourne, Australia \\
$^{3}$ Centre for Eye Research, Melbourne University, East Melbourne, Victoria, Australia \\}

\begin{document}
%
\maketitle
\begin{abstract}
The annotation of disease severity for medical image datasets often relies on collaborative decisions from multiple human graders. The intra-observer variability derived from individual differences always persists in this process, yet the influence is often underestimated. In this paper, we cast the intra-observer variability as an uncertainty problem and incorporate the label uncertainty information as guidance into the disease screening model to improve the final decision. The main idea is dividing the images into simple and hard cases by uncertainty information, and then developing a multi-stream network to deal with different cases separately. Particularly, for hard cases, we strengthen the network's capacity in capturing the correct disease features and resisting the interference of uncertainty. Experiments on a fundus image-based glaucoma screening case study show that the proposed model outperforms several baselines, especially in screening hard cases.
\end{abstract}
\section{Introduction}
Deep learning (DL) models for image-based disease screening heavily rely on large-scale datasets consisting of ($x_{i}, y_{i}$) pairs, where $x_{i}$ is the image data instance and $y_{i}$ is the corresponding label of disease severity. Practically, the label annotation is performed by multiple human experts (e.g., doctors or trained graders) in a collaborative pattern, where the consistent decision from the majority is often regarded as the ``ground truth" of $y_{i}$ \cite{li2018automated,gulshan2016development,li2018efficacy,kermany2018identifying}. Given individual differences in e.g., domain expertise, judgement and bias, intra-observer variability always persists in the annotation process \cite{irvin2019chexpert,abrams1994agreement}. However, defining ground truth by majority consistency is somehow oversimplified, underestimating the influence of intra-observer variability and may even mislead the ground truth. For instance, two images, one is labelled as ``positive'' by $100\%$ of graders while the other one by $51\%$ of graders, will have the same final annotation as ``positive'', despite the grading difficulties being intuitively unequal. 

\begin{figure}\centering
\includegraphics[width=0.48\textwidth]{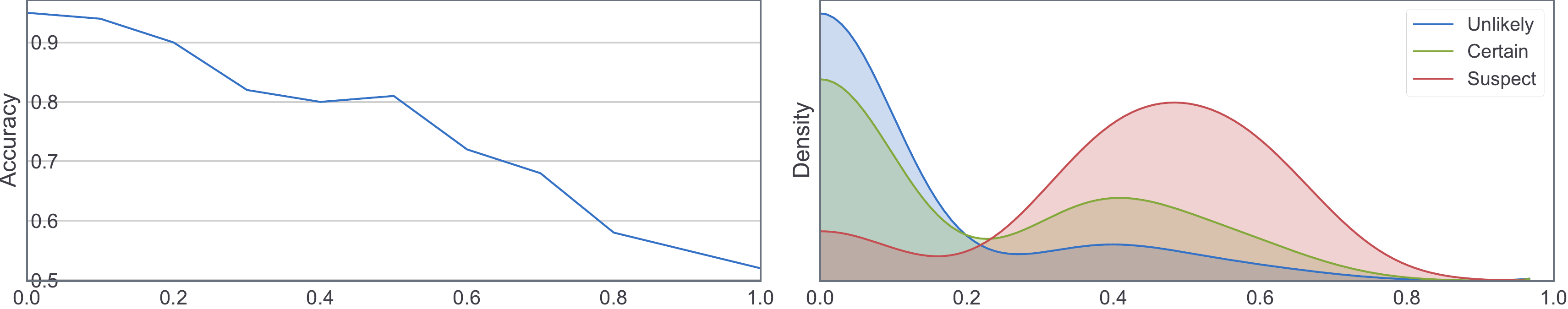}
\caption{(a) A DL model's screening accuracy in different label uncertainty groups; (b) GON severity distributions on different label uncertainty. Both x-axes are label uncertainty scores.} \label{fig1}	
\end{figure}

In fact, the label uncertainty resulted from intra-observer variability contains important prior information, which implies how difficult an image is in identifying its disease severity. Previous studies exploited such information for quantifying imaging quality \cite{liao2019modelling}, or identifying hard patient cases that may require a medical second opinion \cite{raghu2019direct}. Different from previous viewpoints, we believe that label uncertainty can provide prior guidance to improve DL models' decisions, and should be carefully considered in the initial model design. We have two observations in a fundus image-based glaucomatous optic neuropathy (GON) screening task: 

\textit{\textbf{O1:} A DL model generally performs worse on samples with higher label uncertainty.} Fig.\ref{fig1}.a shows the performances of a screening model in different uncertainty score (computed by Eq.\ref{eq1}) groups. The average group accuracy degrades with the uncertainty score increases. 

\textit{\textbf{O2:} Disease severity distribution is potentially correlated with label uncertainty.} Fig.\ref{fig1}.b shows the distributions of GON severity in images with different uncertainty scores. The label uncertainty of suspect GON samples are generally higher than those of unlikely and certain GON samples.

The insights stemming from the observations is that DL screening models should be improved on the samples with higher label uncertainty with identifying ``purer" disease features in the presence of label uncertainty. To this end, we first model label uncertainty as scalar scores using the empirical distribution of intra-observer variability. The images are accordingly divided into simple and hard cases. Then a multi-stream disease screening model is proposed, comprising two main streams, one for simple case screening (SC-Net) and one for hard case screening (HC-Net), and an auxiliary stream (US-Net) for extracting uncertainty-associated features and predicting uncertainty scores. The label uncertainty information is elaborately embedded into the HC-Net's learning process in various ways to offer guidance for classifying hard cases. Specifically, in the training phase, the variability-based encoding and uncertainty-guided joint loss are used to enforce the network to capture correct disease features in the presence of label uncertainty and disentangle the uncertainty-associated features in the latent space. In the inference phase, the uncertainty-guided adaptive threshold is applied to unseen samples to improve the model's decision on hard cases.

\section{Methodology}
\subsection{Modeling label uncertainty}
Given an image $x_{i}$, we let $y_{i}^{(1)}, y_{i}^{(2)}, ..., y_{i}^{(M)}$ be the assigned labels from $M$ individual graders. In practice, the minimum of $M$ is $3$ to gain a significant voting. Letting $c_{1}, ..., c_{K}$ be $K$ classes of disease severity, the empirical distribution of intra-observer variability is $\hat{\mathbf{p}}_{\mathbf{i}}=\left[\hat{p}_{i}^{(1)}, \ldots, \hat{p}_{i}^{(K)}\right]$, with $\hat{p}_{i}^{(k)}=\frac{1}{M} \sum_{m} \mathcal H({y_{i}^{(m)}}=c_{k})$ where $\mathcal H( \cdot) = 1$ when the inside condition stands, otherwise $0$  \cite{raghu2019direct}. Then the label uncertainty of $x_{i}$ can be defined by entropy which indicates the stability of a system:
\begin{equation}
	u_{i} = -\sum_{k} \hat{p}_{i}^{(k)} \log \hat{p}_{i}^{(k)}.
	\label{eq1}
\end{equation}
$u_{i}$ reaches its peak when human disagreements are equally distributed on $K$ classes, indicating the most unstable state.

Note that label uncertainty differs from model uncertainty, also known as epistemic uncertainty. The latter measures the model's confidence in making a decision. Those probabilistic approximation-based method for estimating model uncertainty, such as Monte Carlo Dropout (MC-Dropout) \cite{gal2016dropout}, cannot be directly applied to label uncertainty.

\begin{figure*}\centering
\includegraphics[width=\textwidth]{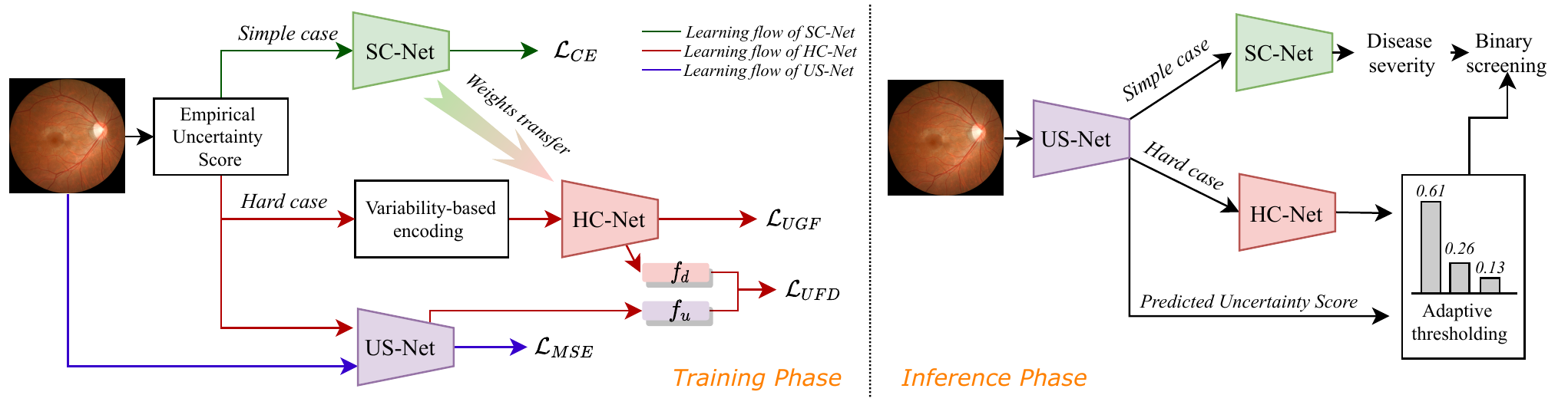}
\caption{Overview of the training phase and the inference phase of the uncertainty-guided multi-stream screening model.} \label{fig2}	
\end{figure*}

\subsection{Multi-stream screening model}
Fig. \ref{fig2} shows an overview of the training phase (left) and the inference phase (right) of the multi-stream screening model. The three sub-streams (US-Net, SC-Net and HC-Net) employ the same deep neural network (DNN) structure as backbones. Notably, the proposed model is independent with DNN structures and compatible with various DNN backbones. 

\subsubsection{US-Net stream}
The US-Net is an auxiliary stream for disentangling uncertainty-associated features and predicting uncertainty scores for unseen samples (see Section \ref{SC-Net and HC-Net streams}). It is optimized with a mean squared error (MSE) loss (Fig.\ref{fig2}: the purple learning flow). Given the training data pair $(x_{i}, u_{i})$, the MSE loss is: 
\begin{equation}
	\mathcal{L}_{MSE} =\frac{1}{N} \sum_{i}\left(u_{i}-\tilde{u}_{i}\right)^{2},
\end{equation}
where $\tilde{u}_{i}$ is the predicted uncertainty score.

\subsubsection{SC-Net and HC-Net streams}
\label{SC-Net and HC-Net streams}
In the training phase, the image samples are divided into simple cases and hard cases via a predefined threshold on the empirical uncertainty score $u$. Then the two cases are fed into the SC-Net and the HC-Net for learning disease severity representation, respectively. The learning of disease severity can be formulated as a multi-classification problem.

Regarding simple cases, the underlying disease severity features can be well captured by a vanilla DNN. Since human graders easily achieve a high consistency for simple cases, we follow previous studies \cite{li2018automated,gulshan2016development,li2018efficacy,kermany2018identifying}using the majority-voting result $\bar y$ as the training ground truth. For the training pair $(x_{i}, \bar y_{i})$, the optimization is performed with a cross-entropy loss (shown as the green learning flow in Fig. \ref{fig2}):
\begin{equation}
	\mathcal{L}_{CE} = - \frac{1}{N} \sum_{i}\sum_{k} \bar p_{i}^{(k)} \log \left(\tilde p_{i}^{(k)} \right),
	\bar{p}_{i}^{(k)} \in \bar{\mathbf{p}}_\mathbf{i}, \tilde{p}_{i}^{(k)} \in \tilde {\mathbf{p}}_\mathbf{i}
\end{equation}
where $\bar{\mathbf{p}}_\mathbf{i}$ denotes the one-hot encoding vector of $\bar y_{i}$ and $\tilde {\mathbf{p}}_\mathbf{i}$ is the class probability yielded by the softmax layer of DNN. 

Compared to simple cases, hard cases are more difficult for both human and vanilla DNNs to recognize the disease severity correctly. Therefore, directly applying the majority-voting-based ground truth and the SC-Net for hard case learning is impractical. Taking advantage of label uncertainty information, we design the HC-Net with several specific strategies to address the problem. The red arrow in Fig. \ref{fig2} illustrates the learning flow of HC-Net.

\textbf{Variability-based encoding.} Rethinking the above one-hot encoding method for a multi-classification problem, where the target class is assigned with a scalar $1$ while other classes are $0$, a strong assumption is that all samples belonging to the same class contain equal amount of ground truth information in the label space, irrespective of their intrinsic difficulties in disease screening being different. This could lead to a model bias in identifying the correct disease features for hard cases. Instead, we propose the variability-based encoding method which applies the empirical distribution of intro-observer variability $\hat{\mathbf{p}}_{\mathbf{i}}=\left[\hat{p}_{i}^{(1)}, \ldots, \hat{p}_{i}^{(K)}\right]$ as the ground-truth. The label uncertainty information retained in the encoded vector can help ensuring the model to capture disease features properly in the presence of uncertainty. Particularly, when all graders arrive at the same decision (i.e., no uncertainty exists), $\hat{\mathbf{p}}_{\mathbf{i}}$ equals to the one-hot vector $\bar{\mathbf{p}}_{\mathbf{i}}$. 

\textbf{Uncertainty-guided focal loss.} Inspired by the focal loss \cite{lin2017focal}, we propose the uncertainty-guided focal loss to replace the cross-entropy loss, which can promote the HC-Net to pay more attentions to hard cases during training: 
\begin{equation}
	\mathcal{L}_{UGF} = - \frac{1}{N} \sum_{i=1}^{N}\sum_{k=1}^{K} \left[ \left(1-\tilde p_{i}^{(k)}\right)^{g(u_{i})} \cdot \hat p_{i}^{(k)} \log \left(\tilde p_{i}^{(k)} \right) \right], \\
\end{equation}
where $\left(1-\tilde p_{i}^{(k)}\right)^{g(u_{i})}$ is imposed to adjust the model's focus according to the difficulty of given samples, i.e., attaching more importance to the samples with lower prediction confidences and higher uncertainty scores; $g(u_{i}) = \gamma \cdot u_{i}$ is an adjusting function relying on $u_{i}$ with a constant weight $\gamma$.


\textbf{Uncertainty feature decoupling loss.} According to the observations in Fig. \ref{fig1}, the potential correlation between uncertainty and disease severity may bias the screen model's decision. Therefore, we propose the uncertainty feature decoupling loss to disentangle disease features $f_{di}$ from uncertainty features $f_{un}$ in the latent space.
\begin{equation}
	\mathcal{L}_{UFD} = - \frac{1}{N} \sum_{i}\max (0, h(u_{i})- \mathcal D(f_{di}(x_{i}), f_{un}(x_{i}))),
\end{equation}
where $f_{di}(x_{i})$ and $f_{un}(x_{i})$  are the flattened feature vectors of image $x_{i}$ output by the last convolutional layer of the HC-Net and the pre-trained US-Net, respectively; $h(u_{i}) = \min (\alpha \cdot u_{i}, 1)$ imposes a $u_{i}$-guided dynamic margin to ensure a lower bound of the two features' distance, where the case is harder, the margin is larger. $\mathcal D(\cdot)$ is the Pearson distance \cite{fulekar2009bioinformatics}.  

Consequently, the final loss for optimizing the HC-Net is:
\begin{equation}
	\mathcal{L}_{joint} = \mathcal L_{UGF} + \mathcal{L}_{UFD}.
\end{equation}

\textbf{Uncertainty-guided adaptive threshold.} The practical population-scale disease screening models are normally expected to make a binary decision in the inference phase to support clinical recommendations, e.g., ``non-referable'' versus ``referable'' cases \cite{li2018automated,gulshan2016development,li2018efficacy,kermany2018identifying}. A common way is applying a fixed threshold (e.g. $0.5$) to $\tilde {\mathbf{p}}_{i}$ to identify the best trade-off between sensitivity and specificity of a classification model \cite{freeman2008comparison,han2018classification,cho2019improving}. The fixed threshold could degrade the flexibility facing cases with varying uncertainty. Instead, we design an uncertainty-guided adaptive threshold for this process. 

Given an unseen sample, first its uncertainty score $\tilde u$ is predicted by the US-Net to decide whether the sample should be allocated to SC-Net or HC-Net. For HC-Net screening, an adaptive threshold $\tau$ is applied to the inference probability of the negative class:
\begin{equation}
	\tau = 1-\frac{K-1}{K}(1-\tilde u)^{\beta},
\end{equation}
where $K$ is the number of GON severity classes and $\beta$ is an adjustable weight; $\tau$ rises when $\tilde u$ increases, meaning that for samples with larger uncertainty, the model is allowed to be more inclined to classify them as ``referable''.   

\section{Experiments}
\subsection{Experimental settings}
\begin{figure}\centering
\includegraphics[width=0.5\textwidth]{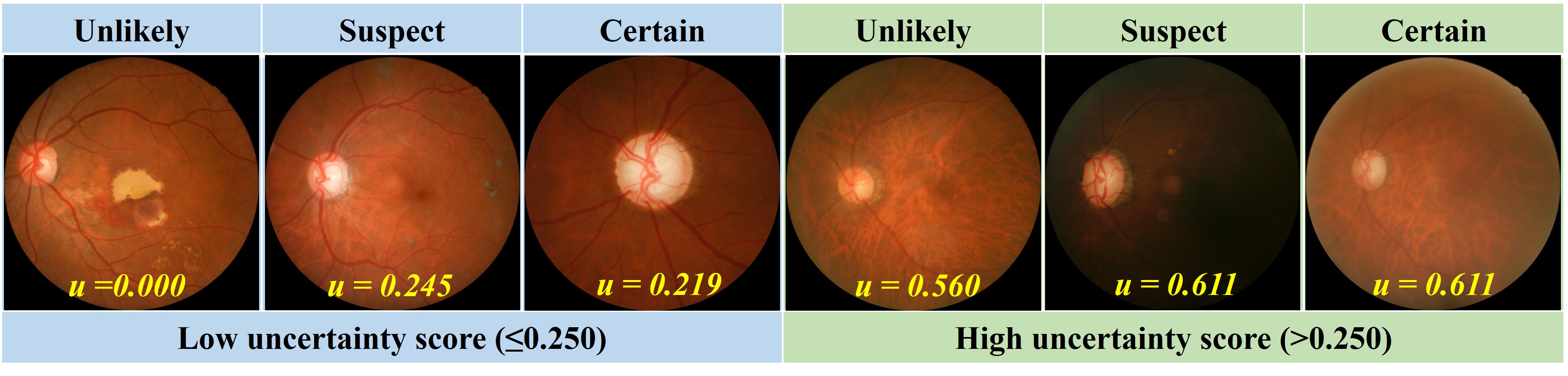}
\caption{Image examples with different GON severity levels and empirical uncertainty scores from the GON dataset.} 
\label{fig3}	
\end{figure}

\noindent \textbf{Datasets.}
The evaluations were performed on the fundus image-based GON screening dataset \emph{LabelMe} \cite{li2018efficacy}. Images were collected from various clinical settings in China. Twenty-one ophthalmologists participated in the grading process. Each image was assigned to different graders sequentially until three consistent individual decisions were achieved. The five-class grading criteria was applied \cite{li2018efficacy}, including unlikely, suspect and certain GON, and poor quality/location. Images with poor quality/location were excluded and the final dataset consists of $47012$ images. The dataset was randomly divided into training, validation and testing sets, as detailed in Table \ref{table1}. Note that in all subsets, the mean uncertainty scores (Eq. \ref{eq1}) of suspect GON samples are significantly higher than the other two classes, in line with our previous observation \textbf{\textit{O2}}. According to the observation in Fig.\ref{fig1}.b, we empirically set the predefined threshold of $u$ as $0.25$. The bottom two rows show the details of simple cases and hard cases divided by this threshold. Fig.\ref{fig3} shows some examples with different GON severity levels and uncertainty scores from this dataset.

\begin{table}[htbp]
  \centering
  \footnotesize
  \caption{The details of the GON dataset. MUS: mean uncertainty score.}
    \setlength{\tabcolsep}{1.7mm}{
    \begin{tabular}{ccccccc}
    \toprule
          & \multicolumn{2}{c}{\footnotesize \textbf{Training set}} & \multicolumn{2}{c}{\footnotesize \textbf{Validation set}} & \multicolumn{2}{c}{\footnotesize \textbf{Testing set}} \\
\cmidrule{2-7}          & \footnotesize \textit{N} & \footnotesize \textit{MUS} & \footnotesize \textit{N} & \footnotesize \textit{MUS} & \footnotesize \textit{N} & \footnotesize \textit{MUS} \\
    \midrule
    \footnotesize \textbf{Unlikely} & 27525 & 0.068  & 6147  & 0.068  & 1511  & 0.067  \\
    \footnotesize \textbf{Suspect} & 2338  & 0.410  & 518   & 0.423  & 94    & 0.412  \\
    \footnotesize \textbf{Certain} & 6947  & 0.162  & 1537  & 0.158  & 395   & 0.151  \\
    \midrule
    \footnotesize \textbf{Simple cases} & 28050 & 0.0003 & 6243  & 0.0003  & 1550  & 0.0003  \\
    \footnotesize \textbf{Hard cases} & 8760  & 0.452  & 1959   & 0.447  & 450    & 0.442  \\
    \bottomrule
    \end{tabular}}%
  \label{table1}%
\end{table}%

\noindent \textbf{Settings.} All images were central-cropped and downsized to $299*299$ with pixel values rescaled to the range of $[0,1]$. We employed Xception \cite{chollet2017xception} as the backbone DNN for all sub-streams. The SC-Net and US-Net were initialized with random weights. The HC-Net were initialized with the pre-trained SC-Net weights in a transfer learning manner \cite{kermany2018identifying}, due to that the shallow layers learn generic low-level domain features that can be shared in similar tasks to accelerate training. The batch size was $220$ and the epoch number was $100$. The Adam optimizer \cite{kingma2014adam} with an initial learning rate of $0.01$ plus an adaptively-decay scheduler was adopted. The hyper-parameters $\{\gamma, \alpha, \beta\}$ were empirically set as $\{4, 1.4, 2\}$ (the decision process is detailed in the supplementary).

\noindent \textbf{Evaluation metrics.} For the multi-classification of GON severity, we computed \textit{F1}-score for each single severity class and the classification accuracy for the overall performance. For the final binary GON screening (i.e., referable vs. non-referable), sensitivity ($SE$), specificity ($SP$) and area under the ROC curve ($AUC$) were evaluated. 

\subsection{Results} 

\noindent \textbf{Ablation experiments.} The ablation experiments were performed in two groups - the whole testing set and only the hard testing cases, respectively. The candidate models include a base model without any strategies (equivalent to the SC-Net) and models with gradually-added strategies. The \textit{uncertainty-guided adaptive threshold} is excluded here since it is only applicable to the final binary screening setting. Table \ref{table2} shows the $F_{1}$ scores of individual classes and the overall classification accuracy. For all classes in both groups, the $F_{1}$ scores and the overall accuracy generally increases with adding more strategies to the base model. The final model combining three strategies outperforms all other models in both groups, and the overall improvement is much more significant in the hard case group ($72.89\%\rightarrow84.22\%$), indicating the effectiveness of the proposed strategies. 

\begin{table}[htbp]
  \centering
  \footnotesize
  \caption{Results of the ablation experiments. \textbf{\textit{M1}}: Base model; \textbf{\textit{M2}}: \textbf{\textit{M1}} + variability-based encoding; \textbf{\textit{M3}}: \textbf{\textit{M2}} + uncertainty-guided focal loss; \textbf{\textit{M4}}: \textbf{\textit{M3}} + uncertainty feature decoupling loss. The highest value is in \textbf{bold}.}
  \setlength{\tabcolsep}{1.1mm}{
    \begin{tabular}{lcccc|cccc}
    \toprule
          & \multicolumn{4}{c|}{\textbf{The whole dataset}} & \multicolumn{4}{c}{\textbf{Hard cases only}} \\
\cmidrule{2-9}          & \scriptsize Unlikely & \scriptsize Suspect & \scriptsize Certain & \scriptsize \textit{Overall } & \scriptsize Unlikely & \scriptsize Suspect & \scriptsize Certain & \scriptsize \textit{Overall} \\
    \midrule
    \scriptsize \textbf{\textit{M1}} & 95.84 & 30.99 & 88.83 & 92.11 & 80.89 & 31.40  & 76.60  & 72.89 \\
    \scriptsize \textbf{\textit{M2}} & 95.83 & 43.04 & 88.70  & 92.28 & 83.13 & 47.41 & 77.42 & 76.01 \\
    \scriptsize \textbf{\textit{M3}} & 96.82 & 55.90  & 90.80  & 93.94 & 85.59 & 58.33 & 79.42 & 79.33 \\
   \scriptsize \textbf{\textit{M4}} & \textbf{97.08} & \textbf{62.35} & \textbf{91.56} & \textbf{94.49} & \textbf{88.70} & \textbf{67.97} & \textbf{85.61} & \textbf{84.22} \\
    \bottomrule
    \end{tabular}}%
  \label{table2}%
\end{table}%

\begin{table}[htbp]
  \centering
  \footnotesize
  \caption{Binary screening performances of the proposed multi-stream model (with three different backbones as shown in brackets) and two baselines. The highest value is in \textbf{bold}.}
  \setlength{\tabcolsep}{1.6mm}{
    \begin{tabular}{lccc|ccc}
    \toprule
          & \multicolumn{3}{c|}{\textbf{The whole dataset}} & \multicolumn{3}{c}{\textbf{Hard cases only}} \\
\cmidrule{2-7}          &\footnotesize $SE$  &\footnotesize $SP$  &\footnotesize $AUC$ &\footnotesize $SE$  &\footnotesize $SP$  &\footnotesize $AUC$ \\
    \midrule
    \footnotesize \textbf{BCNet \cite{li2018efficacy}} & 87.59  & 92.96  & 95.54  & 80.23  & 85.83  & 87.80  \\
    \footnotesize \textbf{DENet \cite{8359118}} & 92.51  & 95.96  & 96.79  & 84.23  & 88.10  & 90.84  \\
    \footnotesize \textbf{CaliNet \cite{jensen2019improving}} & 90.11  & 91.27  & 93.74  & 81.01  & 86.22  & 89.04  \\
    \footnotesize \textbf{Ours (Inception-V3)} & 90.60  & 94.76  & 96.13  & 85.21  & 85.22  & 93.56  \\
    \footnotesize \textbf{Ours (ResNet50)} & \textbf{92.59}  & 96.36 & 98.21  & \textbf{89.44}  & 88.13  & 95.30  \\
    \footnotesize \textbf{Ours (Xception)} & 91.65  & \textbf{97.68}  & \textbf{98.90}  & 89.12  & \textbf{89.57}  & \textbf{95.79}  \\
    \bottomrule
    \end{tabular}}%
  \label{table3}%
\end{table}%

\noindent \textbf{Screening performance.} We tested the final binary GON screening performance of the multi-stream model with all strategies (also including the \textit{uncertainty-guided adaptive threshold}). According to the screening criteria \cite{li2018efficacy}, unlikely GON is regarded as ``non-referable'' while the other two classes are ``referable''. To show our model's compatibility with different DNN structures, we employed three DNNs as backbone, including Inceptivon-V3 \cite{szegedy2016rethinking}, ResNet-50 \cite{he2016deep} and Xception \cite{chollet2017xception}. We also compared our method with two state-of-the-art models for large-scale fundus image-based GON screening: an Inception-V3-based binary classification network (BCNet) \cite{li2018efficacy} and a disc-aware ensemble network (DENet) \cite{8359118}, as well as a label uncertainty-based model calibration method (CaliNet) \cite{jensen2019improving}. Table \ref{table3} shows the results in two groups in terms of $SE, SP$ and $AUC$ scores. The models with three different backbones can all achieve $AUC$ scores larger than $96.00\%$ in the whole dataset group and $93.50\%$ in the hard case group, showcasing its satisfied DNN-compatibility. And their performances are generally better or comparable than the two baselines in both groups. Likewise, the improvement in the hard case group is much more significant, e.g., the $AUC$ score is raised from $90.84\%$ from DENet, which is the best result of baselines, to $95.79\%$ from our method with the Xception backbone. 

\subsection{Conclusion} 
In this paper, we investigated how to leverage the label uncertainty existing in medical image annotation as prior guidance to meliorate disease screening models' decisions. We developed a multi-stream model for the cases with different uncertainty levels, where multiple uncertainty-guided strategies were incorporated specifically for improvement on cases with high label uncertainty. The evaluations conducted in a GON screening case study showed the effectiveness of our method. Our method can benefit general DL models developed on medical image dataset annotated by multiple graders. 

\newpage
\noindent \textbf{Conflicts of Interest.} This work was supported in part by a grant from the Fundamental Research Funds of the State Key Laboratory of Ophthalmology (303060202400362), National Natural Science Foundation of China (grant numbers 81420108008, 81271037, 82101171).
\\
\\
\noindent \textbf{Compliance with Ethical Standards.} This study was performed in line with the principles of the Declaration of Helsinki. Approval was granted by the he institutional review board of Zhongshan Ophthalmic Center (2017KYPJ049). The review board determined that informed consent was not necessary in this study due to the retrospective nature and fully anonymised use of images.
%
%
%
\small
\bibliographystyle{IEEEbib}
\bibliography{mybibliography}

\end{document}